\documentclass[english]{article}
\usepackage[utf8]{inputenc}
\usepackage[T1]{fontenc}
\usepackage{babel}
\usepackage{amsmath}
\usepackage{graphicx}
\usepackage{subcaption}
\usepackage{fancyhdr}
\usepackage{csquotes}
\usepackage{booktabs}
\usepackage{multirow}
\usepackage{xltabular}
\usepackage{longtable}
\usepackage{amstext}
\usepackage{graphicx}
\usepackage[natbib=true,style=numeric,sorting=none]{biblatex}

\begin{document}

\title{Depression Status Estimation by Deep Learning based Hybrid Multi-Modal Fusion Model}

\author{Hrithwik Shalu\textsuperscript{2}, Harikrishnan P\textsuperscript{2}, Hari Sankar CN\textsuperscript{2}, Akash Das\textsuperscript{3},\\ Saptarshi Majumder\textsuperscript{4}, Arnhav Datar\textsuperscript{2}, Subin Mathew MS\textsuperscript{5}, \\ Anugyan Das\textsuperscript{2} and Juned Kadiwala\textsuperscript{1{*}}}

\maketitle
\thispagestyle{fancy}
\noindent
1. University of Cambridge
\\
2. Indian Institute of Technology Madras
\\
3. Indian Institute of Technology Patna
\\
4. Indian Institute of Technology Bombay
\\
5. Jawaharlal Nehru Tropical Botanic Garden and Research Institute
\\
{*}corresponding author

\begin{abstract}
Preliminary detection of mild depression could immensely help in effective treatment of the common mental health disorder. Due to the lack of proper awareness and the ample mix of stigmas and misconceptions present within the society, mental health status estimation has become a truly difficult task. Due to the immense variations in character level traits from person to person, traditional deep learning methods fail to generalize in a real world setting. In our study we aim to create a human allied AI workflow which could efficiently adapt to specific users and effectively perform in real world scenarios. We propose a Hybrid deep learning approach that combines the essence of one shot learning, classical supervised deep learning methods and human allied interactions for adaptation. In order to capture maximum information and make efficient diagnosis video,audio, and text modalities are utilized. Our Hybrid Fusion model achieved a high accuracy of $96.3\%$ on the Dataset; and attained an AUC of $0.9682$ which proves its robustness in discriminating classes in complex real-world scenarios making sure that no cases of mild depression are missed during diagnosis. The proposed method is deployed in a cloud based smartphone application for robust testing. With user specific adaptations and state of the art methodologies, we present a state-of-the-art model with user friendly experience. 

\end{abstract}

\pagebreak

\section*{Introduction}

Depression is considered to be one of the world's leading cause of human disability and in turn increase suicidal tendencies in the community. Close to 800,000 people die due to depression inflicted suicide every year. In the recent years, the rate of depression in successive age cohorts has risen at large. In a survey conducted recently it was found that about 20\% – 30\% of adolescents report symptoms of depression. While depression once was considered an “adult” affliction, the mean age of onset today is 15 i.e, almost 9 percent of high school students have attempted suicide in the past year. With the prevailing pandemic situation, the surge in mental health conditions is tremendous. An individual suffering from depression would show reduced interest and productivity in their day to day activities, leading to a poor overall quality of life. In addition to this depression has been proven to be a core cause of several physical conditions such as tuberculosis or cardiovascular diseases, making it a major health concern around the globe. The most alarming fact is that people around the individual seldom realize the deprived mental state the person is in and sufferers are also more likely to be not forthcoming about how they feel. Having analyzed the above facts it is evident that the depression status estimation of an individual is of utmost importance. Research has proved that facial expression, prosody, syntax, and semantic traits from communications of a person can reflect whether they are in a depressed state of mind, and it is even possible to make out the degree of mental depravity\cite{matanxiety}.  Qualitative speech aspects like speaking rate, speech pauses and voice quality have directly been associated with depressive states \cite{acoustic}\cite{facialexp} , and frequent use of first person singular pronouns\cite{verbal} and a decreased use of complex syntactic constructions such as adverbial clauses\cite{workingmem} are indicators of depression.

For getting more insights to the depression diagnosis done by professionals and to validate our AI-aided preliminary diagnosis approach, we had a discussion with an experienced clinical psychiatrist.  An AI allied approach to diagnose depression and anxiety was deemed extremely helpful, as the method can help in pointing out mild cases which are often missed out by clinical methods. An early stage analysis which can be achieved through our model will act as a tool for tracking the urgent cases and determining the degree of triage.The interaction with the AI system with a patient can be tailored to the specific socio-environmental setting of the doctor, and can be tuned to extract critical information like personality traits and genetic traits without direct consulting approach . The novelty of using class feature comparison methods were established as it was agreed upon that similar depression cases share similar core symptoms which can by captured using multiple modalities. The World Health Organisation(WHO) lists depression as the fourth most significant cause of disability worldwide and predicts it to be the leading cause in 2020. COVID-19 pandemic and the lockdown which followed certainly witnessed a spike in depression and anxiety being diagnosed across the globe. The fact that the number of cases are increasing within all age groups and a resurgence of symptoms in previous patients was inferred from our discussion with the clinical psychiatrist, with working population and children being more severely affected than others.

In our paper we exploit the the proposition that the auditory and visual human communication complement each other, also making use of the text transcripts for more refined diagnosis. We propose a novel multi modal framework comprising of audio-video-text fusion for depression diagnosis. Modes are, essentially, channels of information. These data from multiple sources are semantically correlated, and sometimes provide complementary information to each other, thus reflecting patterns that aren't visible when working with individual modalities on their own. Such systems consolidate heterogeneous, disconnected data from various sensors, thus helping produce more robust predictions. In our method of analysing depression, the video, audio, and text modalities are combined together by using Hybrid Fusion.

\section{Related Work}
With effective sensing being an active part of the research for the last few decades, the active literature that tries to analyze properties of depression with the help of automated means is growing. Chowdhury et al. \cite{rltdwrks} argued that depression constituted a genuine test in individuals and general well-being. They investigated the possibility to utilize social media to try and identify potential signs of significant depression issues in people. Through their web-based social networking postings, they quantified behavioral credits identifying with social engagement, feeling, dialect and semantic styles, sense of the self-system, and notices of antidepressant medications. O’ Dea et al.\cite{ODEA2015183}  in their investigation showed the potential, wherein Twitter can be used as a method for recognizing depression and suicidal tendency in the society. Through human coders and a programmed machine classifier, the authors showed how it is conceivable to recognize the level of worry among suicide-related tweets. While Coppersmith et al. \cite{dist} has shown that modeling of word-use and social language combined with network analysis has been effective in recognizing depression, other researchers exploited sentiment analysis (Xue et al., 2014 \cite{xue2014detecting}; Huang et al., 2014 \cite{inproceedings1}; Yadav et al., 2018a \cite{PB}), topic modeling (Resnik et al., 2015 \cite{resnik-etal-2015-beyond}) and emotion features (Chen et al., 2018 \cite{10.1145/3184558.3191624}) to detect depression. Nguyen et al. \cite{nguyen} showed how machine learning and statistical strategies can be utilized to separate online messages amongst depression. Bachrach et al.\cite{bachrach2012personality} studied how a user’s activity on Facebook identifies with their identity, as measured by the standard Five-Factor Model. They analyzed relationships between user’s identities and the properties of their Facebook profiles. For instance, the size and thickness of their friendship network, number of transferred photographs, and number of occasions went to, number of gathering enrolment, and number of times the user has been tagged in photographs.
With the rich literature available, researchers ( Wang et al. \cite{Wang2013ADD} and Shen et al. \cite{shen2017depression} ) have tried examining various depression-related features and built a multi-modal depressive model to detect the depressed users.

\section{Methods}
The application was build using android studio \cite{studio2017android}. Firebase \cite{stonehem2016google} is used for storage of user data (test history,etc.) and authentication.The application asks the user some questions (as per model's requirement), and records his response(both audio and video)and sends it over to the cloud hosted model using WebRTC \cite{rescorla2013webrtc}. The multi-modal model was trained using a clinical dataset to train and test for the same. Some of the major issues with such datasets were huge class imbalance and non-stationary audio noise features. In the proposed methodology, all of these issues were tackled effectively.This model then provides an inference(from the response of the users) at an https endpoint. The application then collects the data from this endpoint and displays it on the user's screen accordingly. Simultaneously, the results are stored in the firebase storage for future reference. The smartphone application would pave way for improvement of the existing model and provide ease of access to state of the art depression analysis to everyone.A semi-live training scenario was build on the cloud which enables the model to improve over time gradually without intervention.

\subsection{Application Usage Overview}
The android application is the one that acts as a gateway between the user and the model.
The application has an authentication, and thus allowing both regular user(the one who get diagnosed) and clinicians to access the database. Once logged in, the user gets access to all his/her test histories and consultation status. Meanwhile the clinicians receives all his consultation details.Once the user starts taking test, his video and audio is sent over to the cloud which analyses the data and sends over the result to user's phone. The user is then prompted to share the data (if willing)
to improve the model accuracy. The clinicians too can contribute to the model improvement by analysing the result.

\subsubsection{Application Workflow}

\paragraph{  a.  Authentication:}
All users must undergo a registration process in order to access the application services.

The user can register using the 'sign-up' option and provide their details. The two different categories of users are : The regular user (can make use of the application for self-testing) and clinicians (who can the regular users can consult and contribute to model improvement)(Figure 1(b)).

\paragraph{}
Another aspect of the application is to verify the experts on the domain. For this purpose, clinicians are required to upload their registration certificate and are then verified manually(Figure 1(c)).

\pagebreak

\begin{figure}
  \centering
  \begin{subfigure}{.3\linewidth}
    \centering
    \includegraphics[width = \linewidth, height=7cm]{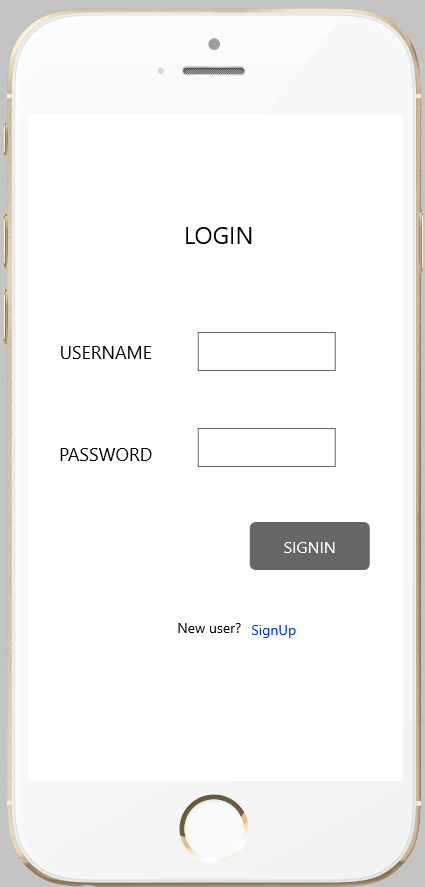}
    \caption{Login page}
  \end{subfigure}
  \hspace{1em}
  \begin{subfigure}{.3\linewidth}
    \centering
    \includegraphics[width = \linewidth, height=7cm]{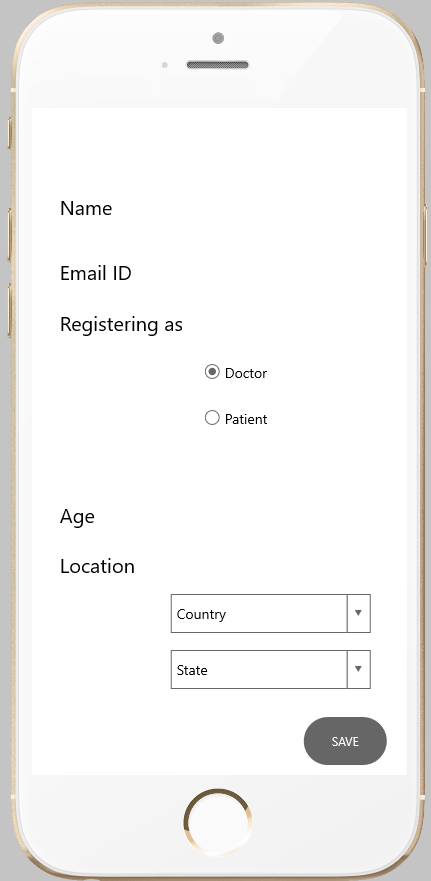}
    \caption{Sign up page}
  \end{subfigure}
  \hspace{1em}
  \begin{subfigure}{.3\linewidth}
    \centering
    \includegraphics[width = \linewidth, height=7cm]{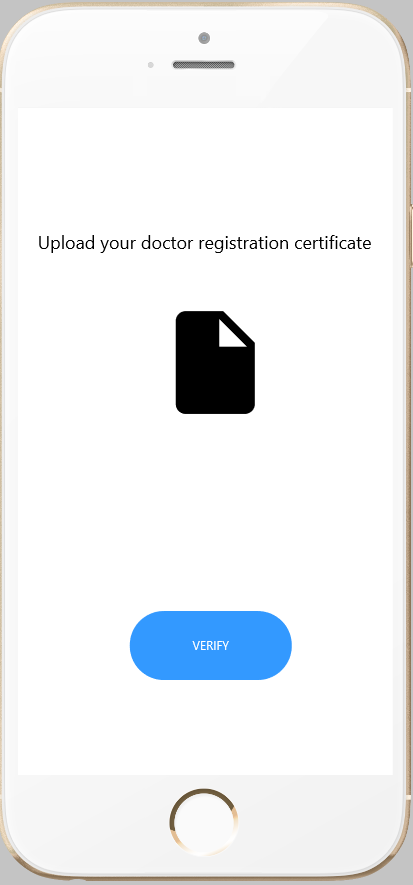}
    \caption{Clinician Verification}
  \end{subfigure}
  \caption{(a) This is the login page for both kind of users.,(b) All the users need to register at this page before logging in.,(c) All the clinicians should be verified and a provision for uploading the required documents is provided here. }
\end{figure}

\paragraph{  b.  Regular user:}
\subparagraph{  b.1  Test History:}
All user details and testing history are stored on the cloud (firebase cloud) and is readily available to the user.The user can also check the consultation status (that is, date and time of consultation) and share their results with  family/clinicians(Figure 2(a)).

\subparagraph{  b.2  Testing:}
Once the user begin to take the test, the application records the video and audio of the user and sends it to the cloud (where the model is hosted) and once the analysis is done it retrieves the result and displays it on user's screen (Figure 2(b)).The questions are either custom (from doctors) or the question from AI system.Also the user can choose to share their test footage and result if they are willing, so that it could be used to benchmark the system and develop it further(Figure 2(c)).

\begin{figure}
  \centering
  \begin{subfigure}{.3\linewidth}
    \centering
    \includegraphics[width = \linewidth, height=7cm]{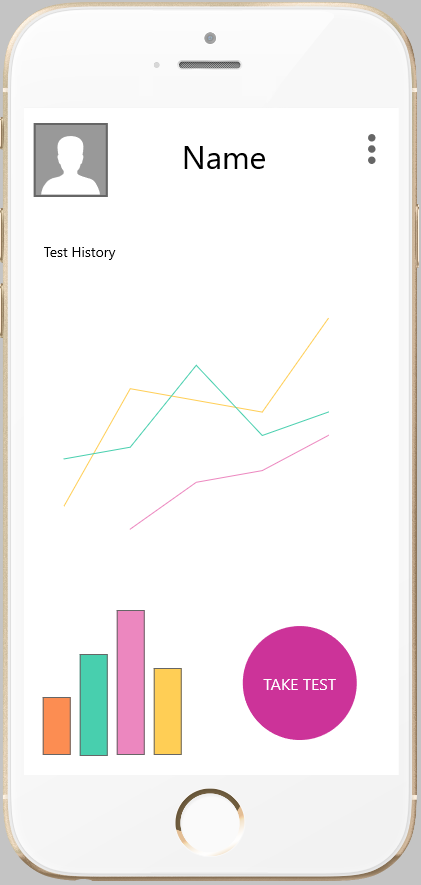}
    \caption{Testing History}
  \end{subfigure}
  \hspace{1em}
  \begin{subfigure}{.3\linewidth}
    \centering
    \includegraphics[width = \linewidth, height=7cm]{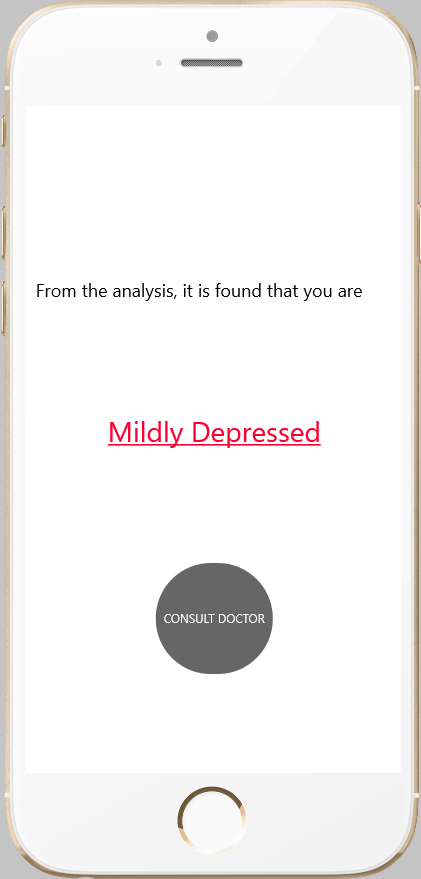}
    \caption{The Result Page}
  \end{subfigure}
  \hspace{1em}
  \begin{subfigure}{.3\linewidth}
    \centering
    \includegraphics[width = \linewidth, height=7cm]{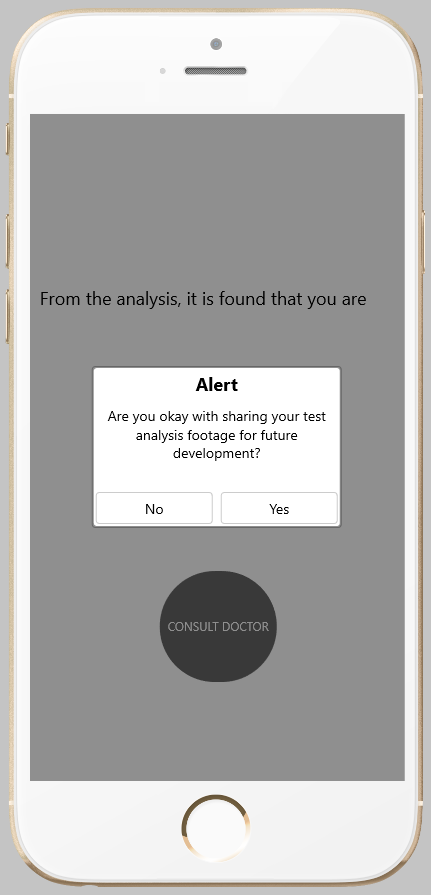}
    \caption{Sharing permission}
  \end{subfigure}
  \caption{(a)The whole history of patient testing history is displayed here.,(b)The result is displayed here.,(c)The prompt that ask for user information sharing }
\end{figure}

\paragraph{  c.  Clinician ( Domain expert ):}
The clinician is given the freedom to set his own questionnaire and also to consult the users according to their depression status and time(Figure 3(b)).Also they can reschedule the consultation timings.

\begin{figure}
  \centering
  \begin{subfigure}{.3\linewidth}
    \centering
    \includegraphics[width = \linewidth, height=7cm]{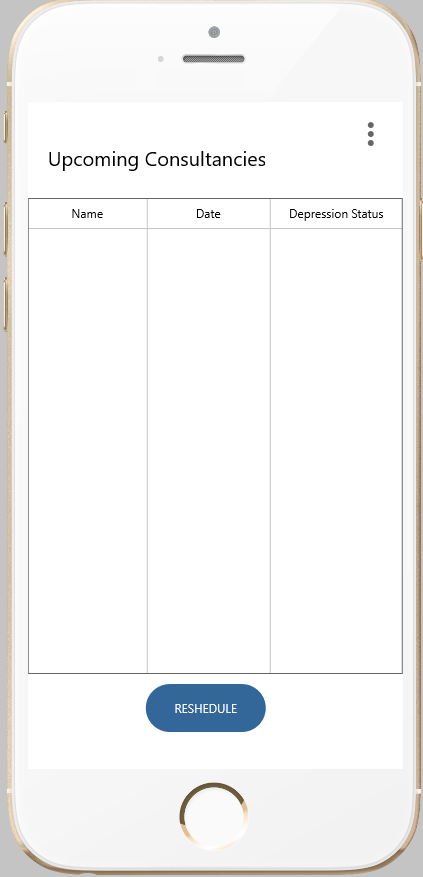}
    \caption{Clinician homepage}
  \end{subfigure}
  \hspace{1em}
  \begin{subfigure}{.3\linewidth}
    \centering
    \includegraphics[width = \linewidth, height=7cm]{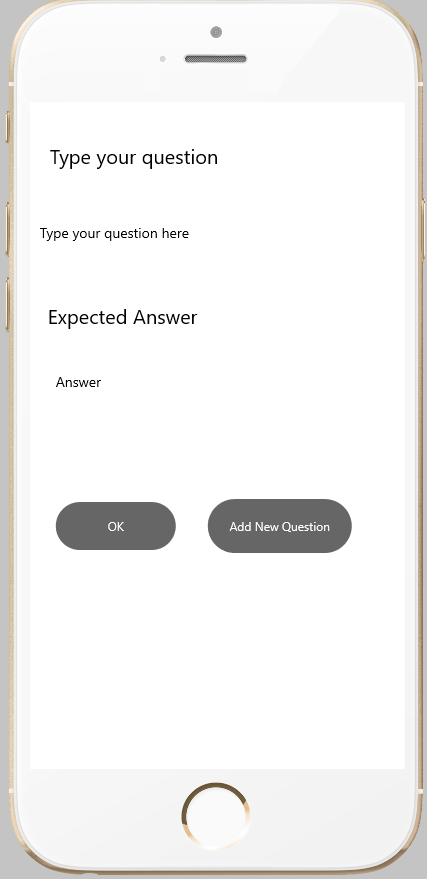}
    \caption{Custom questionnaire}
  \end{subfigure}
  \caption{(a) This shows the login screen of clinician. The entire patient list along with their consultation time and depression status is shown., (b) A custom  questionnaire could be setup by the domain expert.}
\end{figure}

\pagebreak

\subsection{Backend Model}

\begin{figure}[h!]
\centering
\advance\leftskip-4.5cm
\includegraphics[width=21cm, height=14cm]{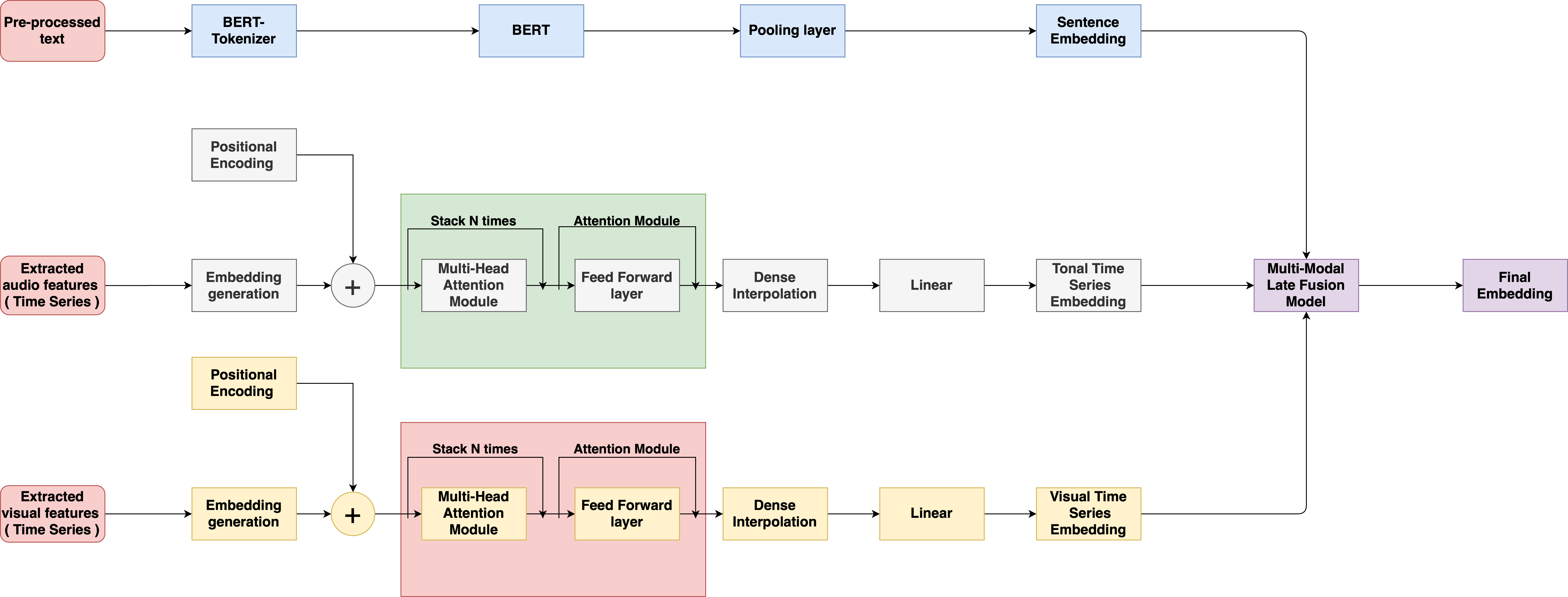}
\caption{The full model architecture used in our study is illustrated}
\label{c2}
\end{figure}

In this section we describe the multi-modal fusion model used for our purposes. The initial task is to pre-process the multi modal data. Then we move on to the feature extraction phase which is carried out using pre-built feature extraction tools. The multi-modal data is made into time-series segments. The obtained segments are then used to train the individual models. The architecture of the full model used in this study is as in Figure \ref{c2}. Pre-processing and training steps taken are detailed below.

\subsubsection{Data Pre-processing}
For initial deployment and robust testing of our model, we train it on the AVEC depression dataset \cite{avecdataset} which fits our overall purpose. Initially  we scrub off the interviewer's intervention in the recording, we then segment the resulting data stream into sequences of 5 minutes each and perform the following preprocessing to the multi-modal time-series segments.

\paragraph{  a.  Text:}
Out of the three modalities text requires least pre-processing. Prior to passage through the  model which utilizes siamese sentence-BERT \cite{reimers2019sentencebert}, rigorous text cleansing is carried out on the text segments formed as a precaution to avoid unforeseen errors.

\paragraph{  b.  Visual:}
To extract features of relevance to our task we require the prominent facial features of an individual, the same is extracted using an open-source behavioral analysis toolkit known as OpenFace \cite{openface}. Out of the wide variety of features extracted using the methodology we utilize the head pose estimate, facial action units and eye-gaze estimate. A every time step a feature vector is formed by concatenating the features of interest.

\paragraph{  c.  Audio:}
Since the audio modality is more prone to noise factors, robustness to non-stationary background noise in audio segments obtained are key for effective analysis. For maximum compatibility with our methodology we used a Cycle GAN based Audio Noise Filter \cite{cyclegan} to carry out stationary/non-stationary noise cancellation from speech. The architecture of the same wasn't varied from the original version \cite{cyclegan}. Filtered audio samples are used for feature extraction by a pre-build acoustic feature extraction application know as COVAREP\cite{coverapp}.

\subsubsection{Model Architecture and Training}
In our methodology we treat the task of audio and visual feature analysis as a multivariate time-series representation task, to efficiently carry out the same using an attention based multivariate time-series analysis methodology popularly know as the (Simply Attend and Diagnose)\cite{song2017attend} architecture. Initially the individual modality models (visual, audio) are pre-trained using a contrastive loss function for initial representation learning. The individual segments of pre-extracted features from the training set are used to carry out the same. The loss function for our purposes is chosen as

\begin{equation}
\mathcal{L}(W(x_i), W(x_j), x_i, x_j) = \frac{1}{2}( 1 - c )D^2 + \frac{1}{2}max\{0 , m - D\}^2
\label{e1}
\end{equation}

\noindent
The margin parameter ($m$) is the desired distance $D$($W(x_1)$,  $W(x_2)$) for a pair of data-points ($x_i$,  $x_j$) belonging to different classes. The distance $D$ between time-series representations ($W(x_1)$,  $W(x_2)$) are estimated as in equation \ref{e2}.

\begin{equation}
D(W(x_i), W(x_j)) = \left\|W(x_i) - W(x_j) \right\|_2
\label{e2}
\end{equation}

\noindent
The binary label $c$ is taken as mentioned in equation \ref{e3}. Where, $y$ is the actual class label.

\begin{equation}
    c = \begin{cases}
        0 \text{, if ($x_i$,  $x_j$) if $y(x_i)$ == $y(x_j)$ } \\
        1 \text{, if ($x_i$,  $x_j$) if $y(x_i)$ != $y(x_j)$ } \text{.}
    \end{cases}
\label{e3}
\end{equation}

\begin{figure}[h!]
\includegraphics[width=12cm, height=11cm]{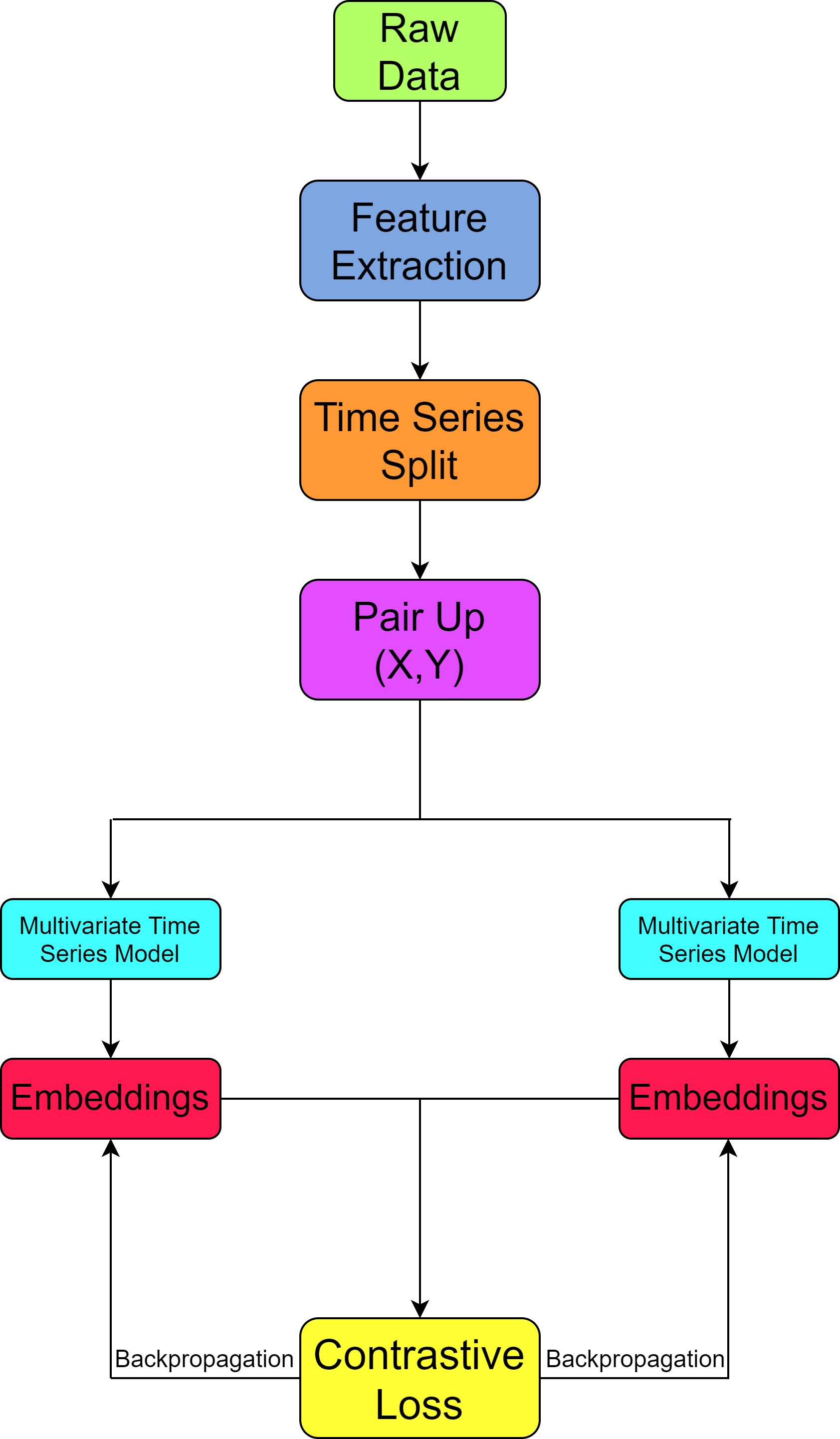}
\caption{Siamese Training Procedure for representation enforcement.}
\label{d21}
\end{figure}

The overview of the training process is illustrated in Figure \ref{d21}. The pre-training in this fashion is crucial to increase the overall efficiency of the fusion model training process. We carry out no initial fine tuning to the pre-trained Sentence-BERT model utilizing RoBERTa \cite{liu2019roberta}.

\begin{figure}[h!]
\includegraphics[width=12cm, height=10cm]{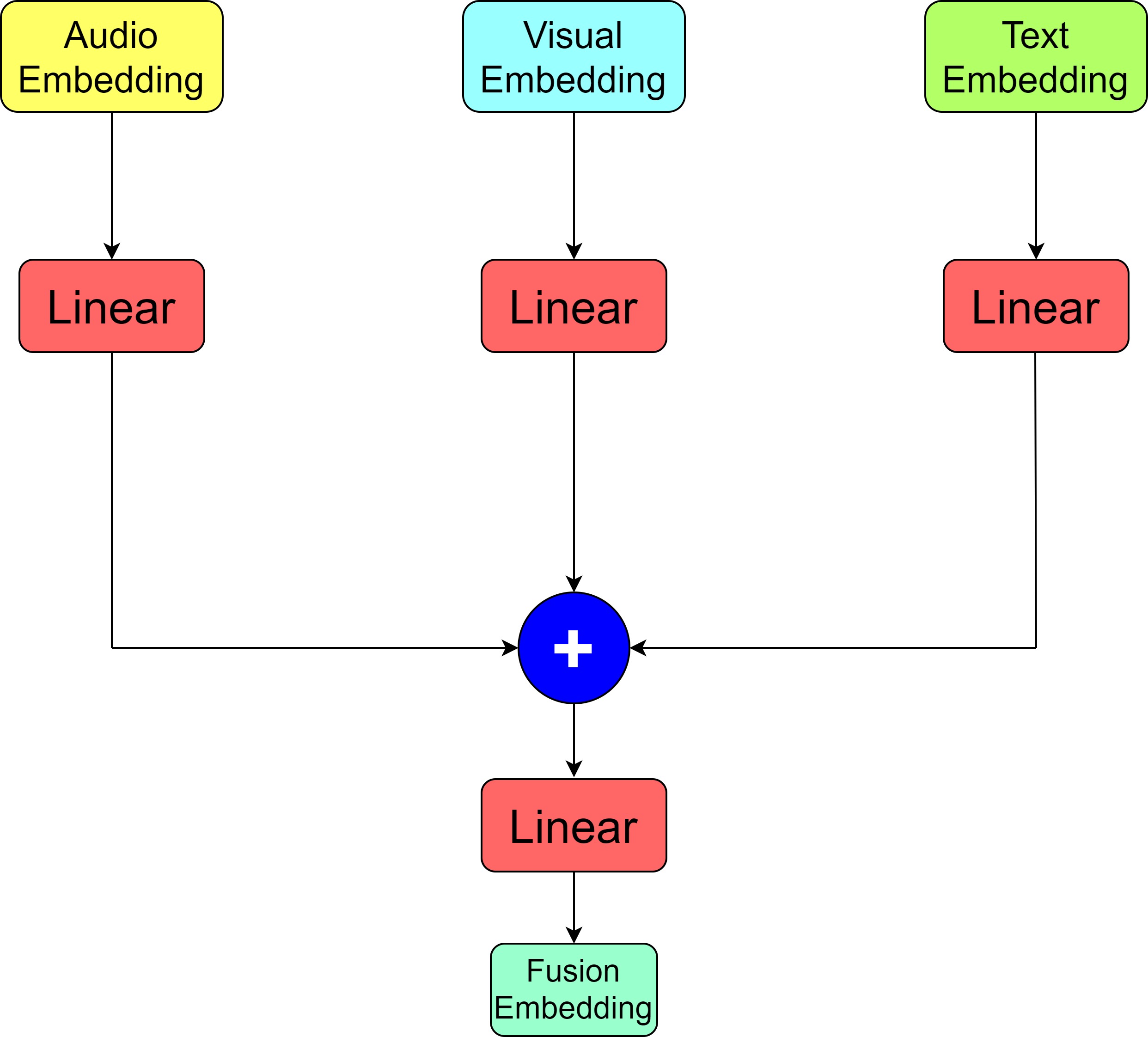}
\caption{Late fusion of multi-modal input}
\label{d22}
\end{figure}

The late fusion of the multi-modal information is one of the core aspects in our entire methodology. The overview of the architecture is as in Figure \ref{d22}. The overall training of the entire model as illustrated in Figure \ref{c2} is carried out by the same loss function as in equation \ref{e1}.

\pagebreak

\subsection{Human Aided Enhancement}
In a real world setting, something as intricate as preliminary depression diagnosis using AI methods must be done with great adaptability so as to suit the specific settings of a clinical expert. Our model comes with the profound inclusion of a semi-live training feature, in which the clinical experts and doctors can keep on adding new patients data to the corpus reinforcing the model's ability to learn. The fact that people belonging to different socio-economic and cultural backgrounds exhibit symptoms of depression in different ways are captured effectively when the corpus is expanded on all fronts. This human aided development greatly improves the model efficiency and predictive power of the methods used as it is exposed to more variety of patient data.

\pagebreak
 
\section{Experiments and Results}

For our purposes we tune our backend model initially by training the same on the publicly available DIAC dataset\cite{avecdataset} ,which is available for academic purposes on reasonable request. We present a comparative analysis, in which the novelty and robustness of the method is established. Dataset description and obtained results are as detailed below. 

\subsection{Dataset}
The dataset used is part of a larger corpus, the Distress Analysis Interview Corpus (DAIC) \cite{avecdataset}, that contains clinical interviews designed to support the diagnosis of psychological distress conditions such as anxiety, depression, and post-traumatic stress disorder. These interviews were collected as part of a larger effort to create a computer agent that interviews people and identifies verbal and nonverbal indicators of mental illness, as a part of study by USC Institute for Creative Technologies. Data collected include audio and video recordings and extensive questionnaire responses;  parts of the corpus have been transcribed and annotated for a variety of verbal and non-verbal features. 

\begin{figure}[h!]
\includegraphics[width=12cm, height=7cm]{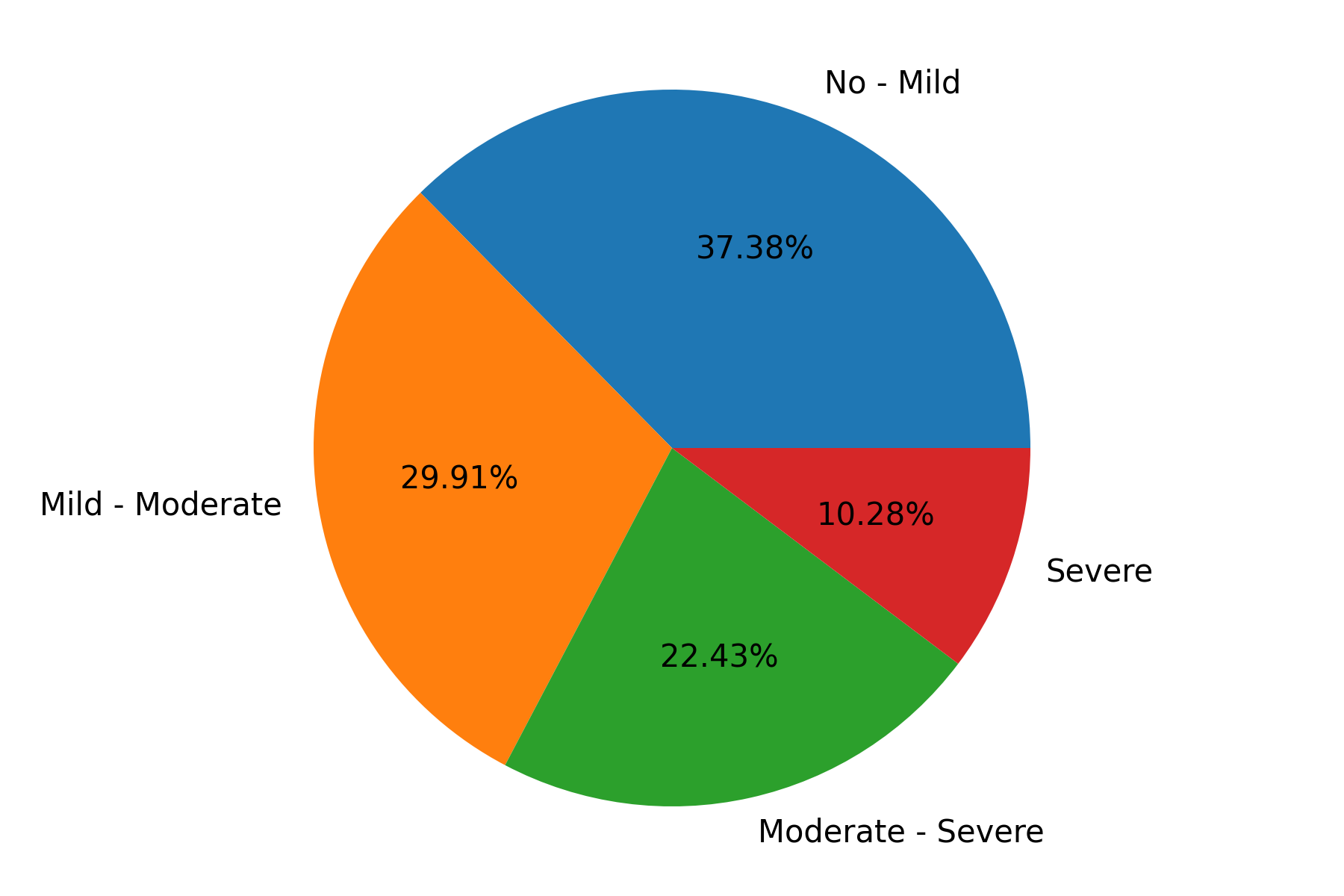}
\caption{Post split class-wise distribution of training data \cite{avecdataset}}
\label{d1}
\end{figure}

\begin{figure}[h!]
\includegraphics[width=12cm, height=6cm]{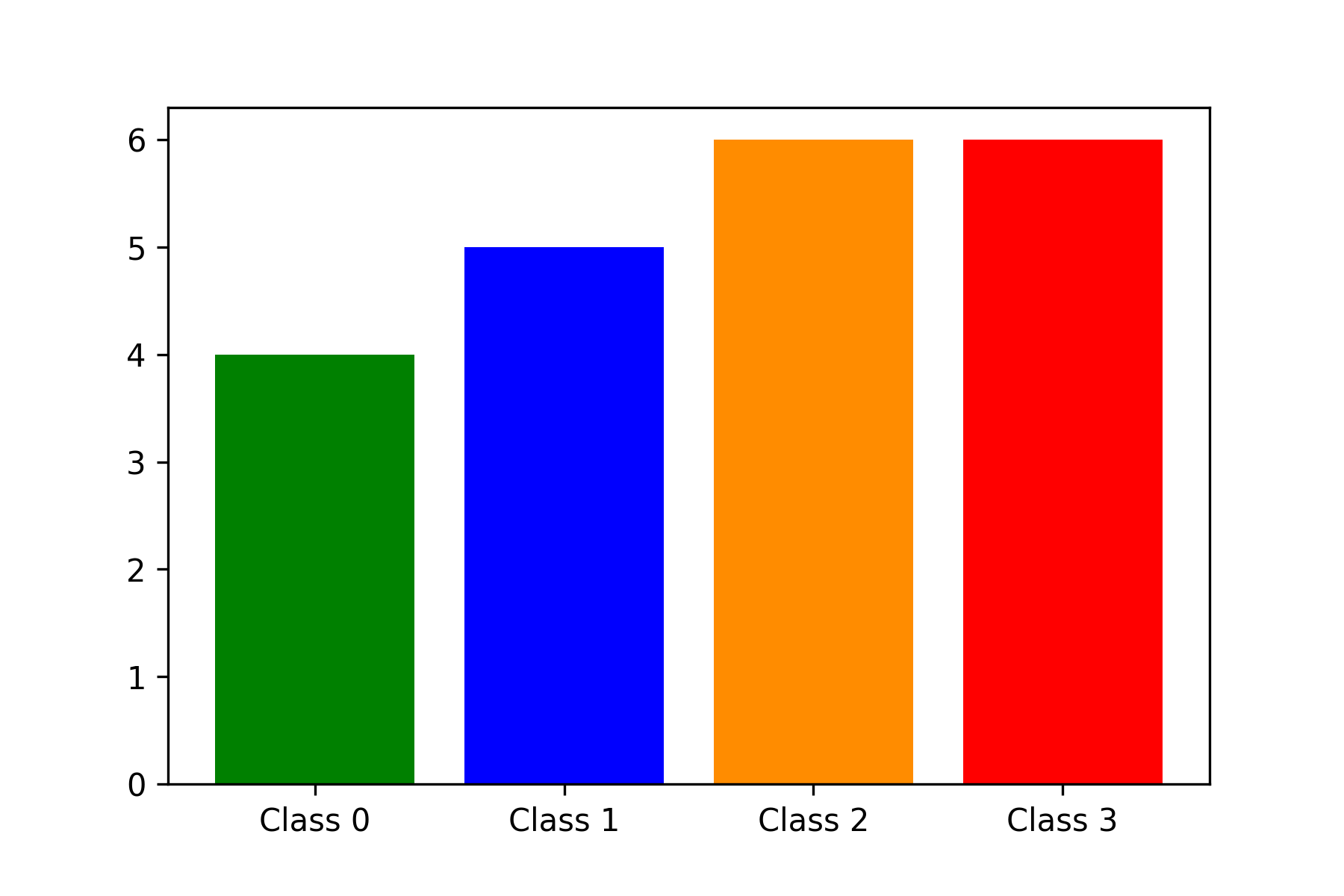}
\caption{Post split distribution of classes\cite{avecdataset}}
\label{d2}
\end{figure}

Each patient data contains the PHQ-8 score that was assigned after a clinical examination. The eight-item Patient Health Questionnaire depression scale (PHQ-8) is established as a valid diagnostic and severity measure for depressive disorders in large clinical studies\cite{phq8}. Figure \ref{d2} shows the distribution of mild, moderate, moderately severe and severe depression cases found among the patients. These scores are calculated using the responses to a set of specific questions which pertain to a diagnostic questionnaire. OpenFace: An open source facial behavior analysis toolkit \cite{openface} was used for extracting the visual features provided with the dataset, which also included confidence tracking and posture. COVAREP, an open-source repository of advanced speech processing algorithms \cite{coverapp} was used to extract the audio features.

For our purposes we formulate the task as a multi-class classification, In order to do so we assign four distinct labels based on the overall PHQ-8 score. This is done with expert opinion

\subsection{Model benchmarks}

In order to robustly benchmark the performance of our model in a real world setting we compare the predictive power of our model for preliminary depression status estimation with that of a clinician in training. A receiver operating characteristic curve was plotted for normalized and efficient comparison. 
Where, the TPR (True Positive Rate or Sensitivity) and the FPR (False Positive Rate) are measured as
\begin{equation}
    TPR = \frac{TP}{TP + FN}
\end{equation}

\begin{equation}
    FPR = \frac{FP}{FP + TN}
\end{equation}

\begin{figure}[h!]
\includegraphics[width=12cm, height=8cm]{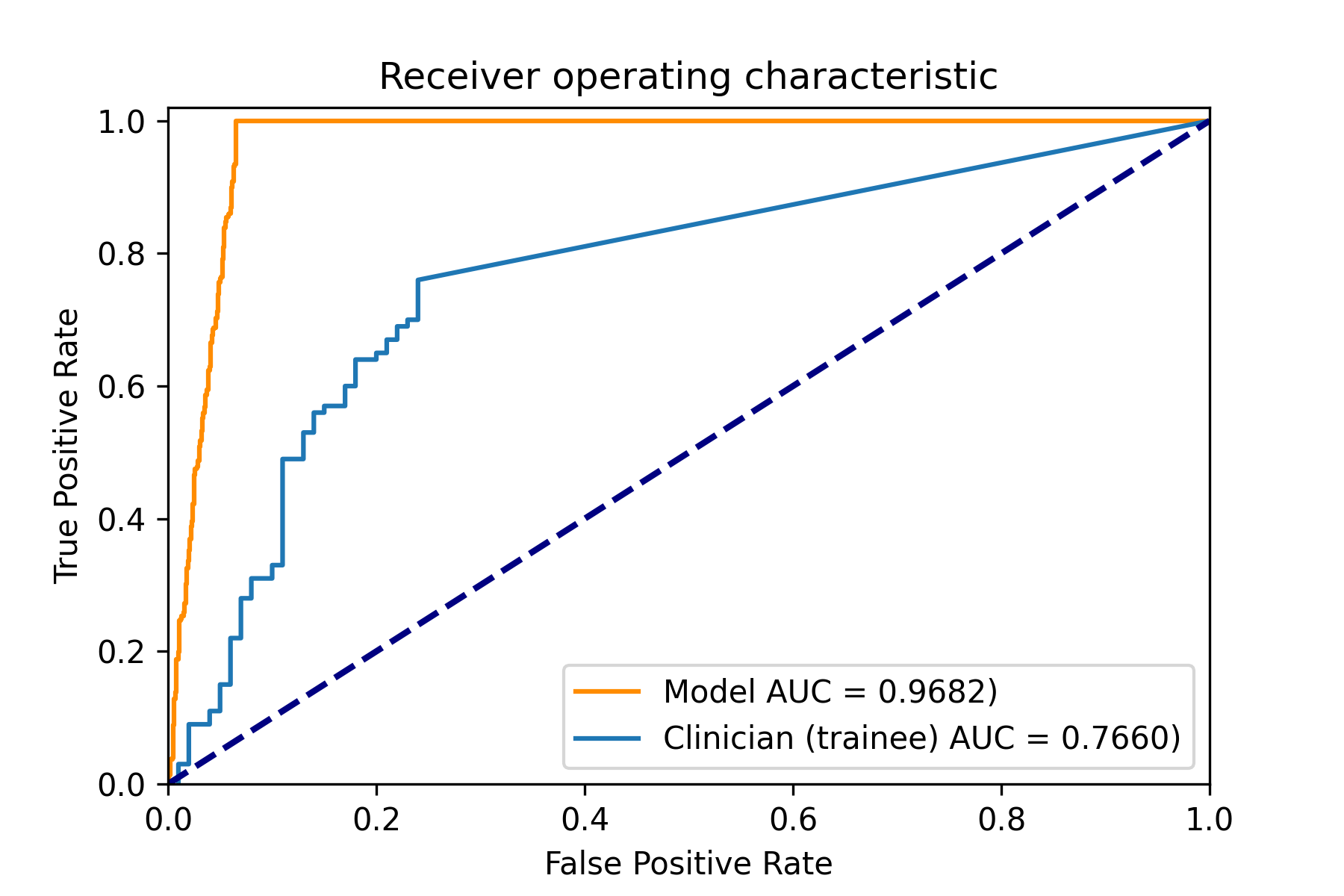}
\caption{ROC curve highlighting real world predictive power}
\label{d3}
\end{figure}

The plot of the experiments are their corresponding AUC values are as in Figure \ref{d3}. The methodology achieved a pinnacle AUC of 0.9682 which in-turn points out the robustness of the same in predicting the right classes in complex real world scenarios.

\pagebreak

\subsection{Comparative Study}

The table showcases some of the standard core questions which are used in evaluating the PHQ-8 score of a patient. The query answers (test) and the standard corpus answers are all given by different patients who exhibited different degrees of depression. They were selected manually to quantify the robustness of our analysis used for preliminary depression status estimation, such that the sets included contextually unique answers from patients belonging to all class labels. Similarity index is a normalized measure of how semantically similar the two candidates in consideration are. Contrasting candidates show similarity indices close to 0 and very similar sequences give similarity index close to 1.

All the labels were predicted correctly by our model, and the values obtained for similarity indices validates the efficiency of our comparative analysis. With indices as high as 0.883 for candidates taken from the same class and as low as 0.128 for those from either ends of the label spectra, it is evident that the model was able to distinguish between various degrees of depression from the interview based data.

\clearpage

\begin{table}[!h]
\advance\leftskip-2.5cm
\resizebox{1.3\textwidth}{!}{%
\begin{tabular}{lllllll}
\hline
\textbf{Questions}                                                                                                                 & \textbf{\begin{tabular}[c]{@{}l@{}}Query Answers \\ ( test )\end{tabular}}                                                                                  & \textbf{\begin{tabular}[c]{@{}l@{}}Label\\ (True)\end{tabular}} & \textbf{\begin{tabular}[c]{@{}l@{}}Label\\ (pred)\end{tabular}} & \textbf{\begin{tabular}[c]{@{}l@{}}Standard  Corpus \\ Answers\end{tabular}}                                                                                                                          & \textbf{Label} & \textbf{\begin{tabular}[c]{@{}l@{}}Similarity\\ Index\end{tabular}} \\ \hline
\multirow{2}{*}{\begin{tabular}[c]{@{}l@{}}How easy is it \\ for you to get a \\ good night's \\ sleep ?\end{tabular}}             & \begin{tabular}[c]{@{}l@{}}i don't have \\ problems sleeping \\ that's i don't ever recall \\ having any problem \\ sleeping so no\end{tabular}             & 0                                                               & 0                                                               & \begin{tabular}[c]{@{}l@{}}uh it's a little difficult \\ but uh of late i've been \\ able to sleep uh within \\ half an hour\end{tabular}                                                             & 1              & 0.2416                                                              \\ \cline{2-7} 
                                                                                                                                   & \begin{tabular}[c]{@{}l@{}}things are going \\ well in my life i generally \\ sleep pretty well\end{tabular}                                                & 0                                                               & 0                                                               & \begin{tabular}[c]{@{}l@{}}well i'm a pretty good \\ sleeper i although  I'm \\ naturally wired to be \\ up later at night so \\ unfortunately my \\ job um allows me\end{tabular}                    & 0              & 0.8823                                                              \\ \hline
\multirow{2}{*}{\begin{tabular}[c]{@{}l@{}}How have you\\ been feeling \\ lately ?\end{tabular}}                                   & \begin{tabular}[c]{@{}l@{}}i guess just uh \\ feeling tired and \\ sluggish and um \\ less less motivated\\  and less interested \\ in things\end{tabular}  & 1                                                               & 1                                                               & \begin{tabular}[c]{@{}l@{}}and up and down like \\ sometimes i've been \\ excited and anxious \\ and on the other side\\  i'm like mm what \\ am i doing\end{tabular}                                 & 1              & 0.7419                                                              \\ \cline{2-7} 
                                                                                                                                   & \begin{tabular}[c]{@{}l@{}}um well i've been \\ i've been depressed \\ for quite some time \\ still having \\ difficulty\end{tabular}                       & 2                                                               & 2                                                               & \begin{tabular}[c]{@{}l@{}}i'm feeling great uh \\ it's an early morning \\ i don't have school\end{tabular}                                                                                          & 0              & 0.1441                                                              \\ \hline
\multirow{2}{*}{\begin{tabular}[c]{@{}l@{}}How are you at \\ controlling your \\ temper ?\end{tabular}}                            & \begin{tabular}[c]{@{}l@{}}i'm alright at it \\ sometimes i lose it \\ i yell when i lose \\ my temper and \\ i have put holes \\ in walls\end{tabular}     & 3                                                               & 3                                                               & \begin{tabular}[c]{@{}l@{}}due to the lack of I \\ guess you could say \\ controlling my \\ emotions i have uh \\ gotten incarcerated \\ sent to prison\end{tabular}                                  & 3              & 0.7256                                                              \\ \cline{2-7} 
                                                                                                                                   & \begin{tabular}[c]{@{}l@{}}not very good \\ seems like \\ everyday i argue \\ with somebody \\ about something \\ but yesterday\end{tabular}                & 2                                                               & 2                                                               & \begin{tabular}[c]{@{}l@{}}good at controlling \\ my temper cuz I \\ don't usually have a\\ temper too often but \\ when i do and i lose \\ my cool i have a \\ hard time controlling it\end{tabular} & 0              & 0.1765                                                              \\ \hline
\multirow{2}{*}{\begin{tabular}[c]{@{}l@{}}What advice \\ would you give \\ yourself ten or\\  twenty years\\  ago ?\end{tabular}} & \begin{tabular}[c]{@{}l@{}}i don't i don't know \\ because i can't say that \\ because um lot of things \\ happened which \\ was not my fault.\end{tabular} & 1                                                               & 1                                                               & to seek help                                                                                                                                                                                          & 3              & 0.1284                                                              \\ \cline{2-7} 
                                                                                                                                   & \begin{tabular}[c]{@{}l@{}}keep working hard \\ and keep thinking \\ about what you do \\ before doing it\end{tabular}                                      & 0                                                               & 0                                                               & \begin{tabular}[c]{@{}l@{}}hmm never to \\ take no for an \\ answer\end{tabular}                                                                                                                      & 0              & 0.7655                                                              \\ \hline
\end{tabular}%
}
\caption{Comparison of candidate answers}
\label{tab:table1}
\end{table}

The tables illustrate the comparative power of the proposed methodology. We believe real world applicability is essential for the success of any study. 
\subsection{Final Model Analysis}

\begin{figure}[h!]
\includegraphics[width=12cm, height=8cm]{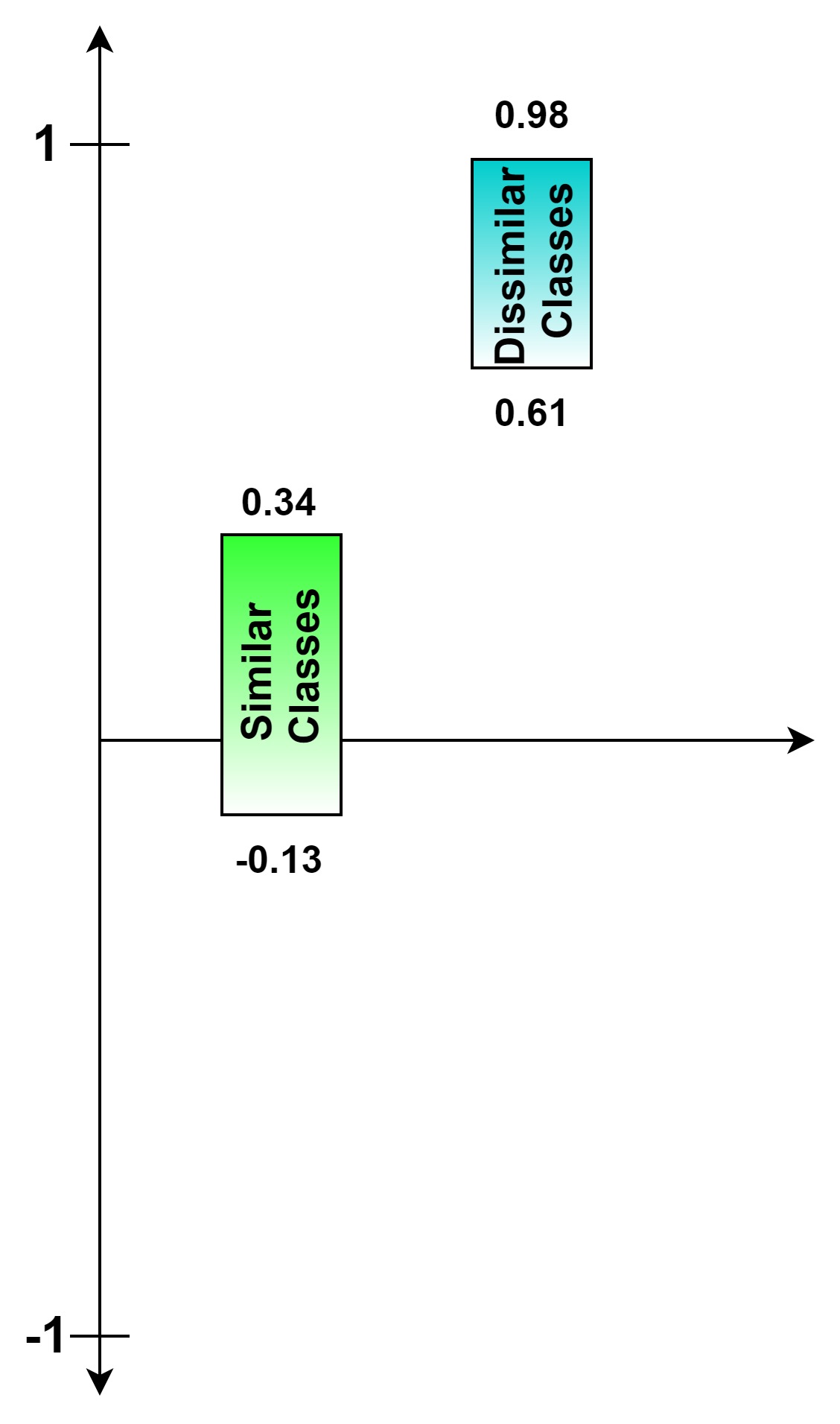}
\caption{Similarity indices class boundary }
\label{r1}
\end{figure}

To conclude with our analysis we test the applicability of our methodology on the test data of the dataset \cite{avecdataset}. We perform comparative analysis between each query candidate belonging to the test set and standard (pre selected with expert opinion) labeled candidates in order to determine the class of each test candidate the model showed amazing performance with a testing accuracy of 96.3\% with a confidence interval of +/- 1.478. The similarity index used for comparison is cosine similarity and the class boundary is highlighted in Figure \ref{r1}. ( Note special cases are excluded ). 

\section*{Conclusion}

Mild depression often goes undetected. Most of the time, people ignore it as a simple day to day issue. But it becomes severe if left untreated, especially in kids(6 - 15 age group) where it might lead to even suicides. This is where the significance of our study comes into picture. The cloud based deployment of the model along with an application ensures the system is easily available and can be used by anyone anywhere in the world. The same provides an effective preliminary analysis for depression diagnosis which can be concluded from the results obtained. Another add-on is the assistance of clinicians on the system which could also be used to adapt and improvise the way in which the system learns. The consultants can even use the system to aid themselves in depression analysis and thus rearrange their consultation order according to the degree of depression. We hope that through the effective deployment of our methodology and it's derivatives, early stage depression conditions could be efficiently detected so that adequate and relevant treatment could be provided.
\section*{Code Availability}
The custom Python code and android app used in this study are available from the corresponding author upon reasonable request and is to be used only for
educational and research purposes.

\printbibliography

@misc{liu2019roberta,
      title={RoBERTa: A Robustly Optimized BERT Pretraining Approach}, 
      author={Yinhan Liu and Myle Ott and Naman Goyal and Jingfei Du and Mandar Joshi and Danqi Chen and Omer Levy and Mike Lewis and Luke Zettlemoyer and Veselin Stoyanov},
      year={2019},
      eprint={1907.11692},
      archivePrefix={arXiv},
      primaryClass={cs.CL}
}

@misc{reimers2019sentencebert,
      title={Sentence-BERT: Sentence Embeddings using Siamese BERT-Networks}, 
      author={Nils Reimers and Iryna Gurevych},
      year={2019},
      eprint={1908.10084},
      archivePrefix={arXiv},
      primaryClass={cs.CL}
}

@inproceedings{avecdataset,
author = {Gratch, Jonathan and Arstein, Ron and Lucas, Gale and Stratou, Giota and Scherer, Stefan and Nazarian, Angela and Wood, Rachel and Boberg, Jill and DeVault, David and Marsella, Stacy and Traum, David and Rizzo, Albert and Morency, L.},
year = {2014},
month = {01},
pages = {},
title = {The Distress Analysis Interview Corpus of human and computer interviews.}
}

@INPROCEEDINGS{openface,
  author={T. {Baltrušaitis} and P. {Robinson} and L. {Morency}},
  booktitle={2016 IEEE Winter Conference on Applications of Computer Vision (WACV)}, 
  title={OpenFace: An open source facial behavior analysis toolkit}, 
  year={2016},
  volume={},
  number={},
  pages={1-10},
  doi={10.1109/WACV.2016.7477553}}

@INPROCEEDINGS{coverapp,  author={G. {Degottex} and J. {Kane} and T. {Drugman} and T. {Raitio} and S. {Scherer}},  booktitle={2014 IEEE International Conference on Acoustics, Speech and Signal Processing (ICASSP)},   title={COVAREP — A collaborative voice analysis repository for speech technologies},   year={2014},  volume={},  number={},  pages={960-964},  doi={10.1109/ICASSP.2014.6853739}}

@INPROCEEDINGS{cyclegan,
  author={N. S. {Nguyen} and T. {Li} and X. {Zhang} and B. {Sheng} and T. {Wang} and J. {Wang}},
  booktitle={2019 IEEE 5th International Conference on Computer and Communications (ICCC)}, 
  title={Audio Noise Filter using Cycle Consistent Adversarial Network - CycleGAN ANF}, 
  year={2019},
  volume={},
  number={},
  pages={884-888},
  doi={10.1109/ICCC47050.2019.9064433}}

@misc{song2017attend,
      title={Attend and Diagnose: Clinical Time Series Analysis using Attention Models}, 
      author={Huan Song and Deepta Rajan and Jayaraman J. Thiagarajan and Andreas Spanias},
      year={2017},
      eprint={1711.03905},
      archivePrefix={arXiv},
      primaryClass={stat.ML}
}

@article{matanxiety,
author = {Morales Santiago and Brown Kayla and Taber-Thomas Bradley and LoBue Vanessa and Buss Kristin and Perez-Edgar Koraly},
year = {2017},
month = {08},
pages = {874-883},
title = {Maternal Anxiety Predicts Attentional Bias Towards Threat in Infancy},
volume = {17},
journal = {Emotion},
doi = {10.1037/emo0000275}
}

@article{acoustic,
author ={Mundt JC and Snyder PJ and Cannizzaro MS and Chappie K and Geralts DS.},
year={2007},
title={Voice acoustic measures of depression severity and treatment response collected via interactive voice response (IVR) technology.},
doi={10.1016/j.jneuroling.2006.04.001}
}

@ARTICLE{facialexp,
 AUTHOR={Scherer, Klaus R. and Ellgring, Heiner and Dieckmann, Anja and Unfried, Matthias and Mortillaro, Marcello},   
 TITLE={Dynamic Facial Expression of Emotion and Observer Inference}, 	 
 JOURNAL={Frontiers in Psychology}, 	 
 VOLUME={10}, 	 
PAGES={508},	 
 YEAR={2019}, 	 
 URL={https://www.frontiersin.org/article/10.3389/fpsyg.2019.00508},  	 
 DOI={10.3389/fpsyg.2019.00508}, 	 
 ISSN={1664-1078},   
}

@article{verbal,
author = {Rude, Stephanie and Gortner, Eva-Maria and Pennebaker, James},
year = {2004},
month = {12},
pages = {1121-1133},
title = {Language Use of Depressed and Depression-Vulnerable College Students},
volume = {18},
journal = {Cognition \& Emotion - COGNITION EMOTION},
doi = {10.1080/02699930441000030}
}

@article{workingmem,
author = {Zinke, Katharina and Zeintl, Melanie and Rose, Nathan and Putzmann, Julia and Pydde, Andrea and Kliegel, Matthias},
year = {2013},
month = {05},
pages = {},
title = {Working Memory Training and Transfer in Older Adults: Effects of Age, Baseline Performance, and Training Gains},
volume = {50},
journal = {Developmental psychology},
doi = {10.1037/a0032982}
}

@inproceedings{rltdwrks,
  title={Predicting Depression via Social Media},
  author={Munmun De Choudhury and M. Gamon and S. Counts and E. Horvitz},
  booktitle={ICWSM},
  year={2013}
}

@inproceedings{dist,
author = {Coppersmith, Glen and Dredze, Mark and Harman, Craig},
year = {2014},
month = {01},
pages = {51-60},
title = {Quantifying Mental Health Signals in Twitter},
doi = {10.3115/v1/W14-3207}
}

@inproceedings{inproceedings1,
author = {Huang, Xiaolei and Zhang, Lei and Liu, Tianli and Chiu, David and Zhu, Tingshao and Li, Xin},
year = {2014},
month = {11},
pages = {},
title = {Detecting Suicidal Ideation in Chinese Microblogs with Psychological Lexicons},
doi = {10.1109/UIC-ATC-ScalCom.2014.48}
}

@inproceedings{PB,
    title = "Medical Sentiment Analysis using Social Media: Towards building a Patient Assisted System",
    author = "Yadav, Shweta  and
      Ekbal, Asif  and
      Saha, Sriparna  and
      Bhattacharyya, Pushpak",
    booktitle = "Proceedings of the Eleventh International Conference on Language Resources and Evaluation ({LREC} 2018)",
    month = may,
    year = "2018",
    address = "Miyazaki, Japan",
    publisher = "European Language Resources Association (ELRA)",
    url = "https://www.aclweb.org/anthology/L18-1442",
}

@inproceedings{resnik-etal-2015-beyond,
    title = "Beyond {LDA}: Exploring Supervised Topic Modeling for Depression-Related Language in {T}witter",
    author = "Resnik, Philip  and
      Armstrong, William  and
      Claudino, Leonardo  and
      Nguyen, Thang  and
      Nguyen, Viet-An  and
      Boyd-Graber, Jordan",
    booktitle = "Proceedings of the 2nd Workshop on Computational Linguistics and Clinical Psychology: From Linguistic Signal to Clinical Reality",
    month = jun # " 5",
    year = "2015",
    address = "Denver, Colorado",
    publisher = "Association for Computational Linguistics",
    url = "https://www.aclweb.org/anthology/W15-1212",
    doi = "10.3115/v1/W15-1212",
    pages = "99--107",
}

@article{nguyen,
author = {Nguyen, Thin and Phung, Dinh and Dao, Bo and Venkatesh, Svetha and Berk, Michael},
year = {2014},
month = {07},
pages = {217-226},
title = {Affective and Content Analysis of Online Depression Communities},
volume = {5},
journal = {IEEE Transactions on Affective Computing},
doi = {10.1109/TAFFC.2014.2315623}
}

@inproceedings{Wang2013ADD,
  title={A Depression Detection Model Based on Sentiment Analysis in Micro-blog Social Network},
  author={X. Wang and C. Zhang and Y. Ji and Li Sun and Leijia Wu and Zhana Bao},
  booktitle={PAKDD Workshops},
  year={2013}
}

@article{phq8,
author = {Kroenke, Kurt and Strine, Tara and Spitzer, Robert and Williams, Janet and Berry, Joyce and Mokdad, Ali},
year = {2008},
month = {09},
pages = {163-73},
title = {The PHQ-8 as a Measure of Current Depression in the General Population},
volume = {114},
journal = {Journal of affective disorders},
doi = {10.1016/j.jad.2008.06.026}
}

@article{rescorla2013webrtc,
  title={WebRTC security architecture},
  author={Rescorla, Eric},
  journal={Work in Progress},
  year={2013}
}

@book{stonehem2016google,
  title={Google Android Firebase: Learning the Basics},
  author={Stonehem, Bill},
  volume={1},
  year={2016},
  publisher={First Rank Publishing}
}

@article{studio2017android,
  title={Android Studio},
  author={Studio, Android},
  journal={The Official IDE for Android},
  year={2017}
}

@article{ODEA2015183,
title = "Detecting suicidality on Twitter",
journal = "Internet Interventions",
volume = "2",
number = "2",
pages = "183 - 188",
year = "2015",
issn = "2214-7829",
doi = "https://doi.org/10.1016/j.invent.2015.03.005",
url = "http://www.sciencedirect.com/science/article/pii/S2214782915000160",
author = "Bridianne O'Dea and Stephen Wan and Philip J. Batterham and Alison L. Calear and Cecile Paris and Helen Christensen"
}

@inproceedings{10.1145/3184558.3191624,
author = {Chen, Xuetong and Sykora, Martin D. and Jackson, Thomas W. and Elayan, Suzanne},
title = {What about Mood Swings: Identifying Depression on Twitter with Temporal Measures of Emotions},
year = {2018},
isbn = {9781450356404},
publisher = {International World Wide Web Conferences Steering Committee},
address = {Republic and Canton of Geneva, CHE},
url = {https://doi.org/10.1145/3184558.3191624},
doi = {10.1145/3184558.3191624},
booktitle = {Companion Proceedings of the The Web Conference 2018},
pages = {1653–1660},
numpages = {8},
location = {Lyon, France},
series = {WWW '18}
}

@inproceedings{xue2014detecting,
  title={Detecting adolescent psychological pressures from micro-blog},
  author={Xue, Yuanyuan and Li, Qi and Jin, Li and Feng, Ling and Clifton, David A and Clifford, Gari D},
  booktitle={International Conference on Health Information Science},
  pages={83--94},
  year={2014},
  organization={Springer}
}

@inproceedings{bachrach2012personality,
  title={Personality and patterns of Facebook usage},
  author={Bachrach, Yoram and Kosinski, Michal and Graepel, Thore and Kohli, Pushmeet and Stillwell, David},
  booktitle={Proceedings of the 4th annual ACM web science conference},
  pages={24--32},
  year={2012}
}

@inproceedings{shen2017depression,
  title={Depression Detection via Harvesting Social Media: A Multimodal Dictionary Learning Solution.},
  author={Shen, Guangyao and Jia, Jia and Nie, Liqiang and Feng, Fuli and Zhang, Cunjun and Hu, Tianrui and Chua, Tat-Seng and Zhu, Wenwu},
  booktitle={IJCAI},
  pages={3838--3844},
  year={2017}
}
\end{document}